\documentclass[conference]{IEEEtran}
\IEEEoverridecommandlockouts
\usepackage[english]{babel}
\usepackage{cite}
\usepackage{amsmath,amssymb,amsfonts}
\usepackage{algorithmic}
\usepackage{graphicx}
\usepackage{textcomp}
\usepackage{xcolor}
\usepackage{xspace}
\usepackage{soul}

\usepackage[spaces,hyphens]{xurl} 
\usepackage[colorlinks,allcolors=blue]{hyperref}

\usepackage{tikz}
\usetikzlibrary{arrows}

\definecolor{codegreen}{rgb}{0,0.6,0}
\definecolor{codegray}{rgb}{0.5,0.5,0.5}
\definecolor{codewhite}{rgb}{1.0,1.0,1.0}
\definecolor{codeorange}{rgb}{0.91,0.45,0.20}
\definecolor{codeblue}{rgb}{0.18,0.55,0.85}
\definecolor{codebluish}{rgb}{0.1,0.6,0.6}
\definecolor{codered}{rgb}{1,0,0}

\usepackage{listings}
\makeatletter
\newcommand{\linebreakand}{%
  \end{@IEEEauthorhalign}
  \hfill\mbox{}\par
  \mbox{}\hfill\begin{@IEEEauthorhalign}
}
\makeatother

\def\BibTeX{{\rm B\kern-.05em{\sc i\kern-.025em b}\kern-.08em
    T\kern-.1667em\lower.7ex\hbox{E}\kern-.125emX}}

\def\FLib{Flotta\xspace}
\def\ImportCode{\lstinputlisting[float=hbt!,caption={Example of Artifact submission.},label=code:example]{code_example.py}}
\def\GitHub{\url{https://github.com/IDSIA/flotta}\xspace}
\def\Docs{\url{https://readthedocs.org/projects/flotta/}\xspace}
\def\Funded{{This work was supported in part by Innosuisse through the Innosuisse Flagship project SPEARHEAD (PFFS-21-15), and by the Swiss State Secretariat for Education, Research and Innovation (SERI)  through the Horizon Europe Project NextGen, which is also funded by the European Union’s Horizon Europe, under grant agreement number 101136962, and by the UK Research and Innovation (UKRI) under the UK government’s Horizon Europe funding guarantee (grant numbers 10104323 and 10098097).}
}

\newif\ifproofread

\begin{document}
\proofreadtrue

\title{
    \textit{\FLib}: a Secure and Flexible Spark-inspired Federated Learning Framework
    \thanks{\Funded}
}
\author{
    \IEEEauthorblockN{1\textsuperscript{st} Claudio Bonesana}
    \textit{IDSIA (USI-SUPSI)}\\
        Lugano, Switzerland \\
        claudio.bonesana@idsia.ch
    
    \and\IEEEauthorblockN{2\textsuperscript{nd} Daniele Malpetti}
    \IEEEauthorblockA{
    \textit{IDSIA (USI-SUPSI)}\\
        Lugano, Switzerland \\
        daniele.malpetti@idsia.ch
    }
    \and
    \IEEEauthorblockN{3\textsuperscript{rd} Sandra Mitrovi\'c}
    \IEEEauthorblockA{
    \textit{IDSIA (USI-SUPSI)}\\
        Lugano, Switzerland \\
        sandra.mitrovic@idsia.ch
    }
    \linebreakand
    \IEEEauthorblockN{4\textsuperscript{th} Francesca Mangili}
    \IEEEauthorblockA{
    \textit{IDSIA (USI-SUPSI)}\\
        Lugano, Switzerland \\
        francesca.mangili@idsia.ch
    }
    \and
    \IEEEauthorblockN{5\textsuperscript{th} Laura Azzimonti}
    \IEEEauthorblockA{
    \textit{IDSIA (USI-SUPSI)}\\
        Lugano, Switzerland \\
        laura.azzimonti@idsia.ch
    }
}

\maketitle

\begin{abstract}
We present \FLib
\footnote{`Flotta' is the Italian word for `fleet', i.e.,  a set of heterogeneous entities acting collaboratively.}, a Federated Learning framework designed to train machine learning models on sensitive data distributed across a multi-party consortium conducting research in contexts requiring high levels of security, such as the biomedical field. \FLib is a Python package, inspired in several aspects by Apache Spark, which provides both flexibility and security and allows conducting research using solely machines internal to the consortium. In this paper, we describe the main components of the framework together with a practical use case to illustrate the framework's capabilities and highlight its security, flexibility and user-friendliness.
\end{abstract}


\section{Introduction}
\label{sec:intro}

Federated Learning (FL) \cite{pmlr-v54-mcmahan17a} allows multiple devices or institutions to collaboratively train a model while keeping the data decentralized. Each party performs training steps locally and independently, without disclosing the data, and shared only (some of) the model parameters for the creation of a single global model. FL emerged as a response to mitigate data privacy and security issues in scenarios with strict legal regulations regarding the manipulation of privacy-sensitive data, such as EU law on General Data Protection Regulation - GDPR \cite{voigt2017eu}. 

FL allows for a large variety of choices regarding different aspects. For instance, FL could be applied both to scenarios with only few parties (commonly evidenced in the healthcare domain) and scenarios with as much as millions of parties (typical for edge devices, such as mobile phones). The literature refers to the former as \textit{cross-silo} \cite{rieke2020future} and to the latter as \textit{cross-device} \cite{bonawitz2019towards}. {In a cross-silo context, e.g., where data are distributed across hospitals,  scalability issues commonly associated with FL are less critical because the number of nodes is relatively small, and they typically have reliable connections and substantial computational capacity. } Additionally, when it comes to the parties' interactions, FL can follow either a \textit{centralized} topology, where one party takes an exclusive role of coordinating the training process, or, the opposite \textit{decentralized} topology, where no dedicated coordinator is present.

{Over the years, several frameworks have emerged to manage FL. Each framework is characterized by distinct features and often relies on different technologies. Some of them are general-purpose, while others focus on specific classes of algorithms, scenarios, topologies or domains of use. For instance, TensorFlow Federated \cite{tff2021}, and PySyft \cite{ziller2021pysyft}) focus on deep learning, Felicitas \cite{zhang2022felicitas}, and FairFed \cite{ur2020fairfed} are frameworks developed with a focus on the \textit{cross-device} scenario, while Substra \cite{galtier2019substra} implements the \textit{decentralized} topology. Among general purpose frameworks we can mention Flower \cite{beutel2022flower},  FedML \cite{he2020fedml}, and FATE \cite{liu2021fate}, while FeatureCloud \cite{matschinske2023featurecloud}, and Sfkit \cite{mendelsohn2023sfkit} are examples of frameworks specifically developed for the biomedical sector that come with user-friendly web interfaces, allowing for practical creation and management of FL consortia. }

{In this paper, we are considering a cross-silo scenario including a relatively small number of parties that wish to collaboratively learn from the sensitive data they own, which is a typical scenario in the healthcare field. While this alleviates scalability concerns, it requires to guarantee high levels of security. In particular we consider a situation where not only data cannot be moved from their original locations, but also  the parties possessing data cannot be exposed to the world. In addition to these requirements, we are also interested in going beyond the classic centralized topology and having a fair balance between flexibility and security when it comes to the code execution.} Despite the plethora of existing FL frameworks, which excel in many different aspects, to the best of our knowledge, there are no existing frameworks which satisfy all the requested criteria. 

We propose \FLib, a framework to bridge this gap, enabling FL in contexts requiring high levels of security, where communication  must remain inside a consortium of authorized partners, while jointly adapting and extending the best practices proposed in existing FL frameworks. For example, \FLib naturally supports both \textit{centralized} and \textit{decentralized} topologies, but it also allows for intermediate scenarios. Additionally, not only \FLib guarantees security, but it also enables a good trade-off between flexibility and security, as the flow of custom analysis pipelines can be built as a combination of pre-defined, granted instructions.
The user is therefore allowed to combine and execute only recognized pieces of code, preventing a malicious user from unveiling sensitive data. This is enabled by the software architecture of \FLib, which has been heavily inspired by Apache Spark \cite{zaharia2016apache}, a unified engine for large-scale data analytics in cluster of computers that permits the transfer of instructions across nodes through predefined operations. 
{While Spark is not directly used in the framework, its approach to distributed computed has inspired our work for what concerns the deployment procedure, and the concept of Workbench, and Artifact detailed in section \ref{sec:framework}.}

The paper is organised as follows: Section~\ref{sec:framework} motivates the development of a new framework; Section~\ref{sec:library} elucidates the principal components of \FLib, while Section~\ref{sec:usage} elaborates on its usage. Finally, Section~\ref{sec:conclusion} concludes providing directions for future works.

\section{Why another framework}
\label{sec:framework}

\begin{figure*}
    \centering
    \includegraphics[width=1\textwidth]{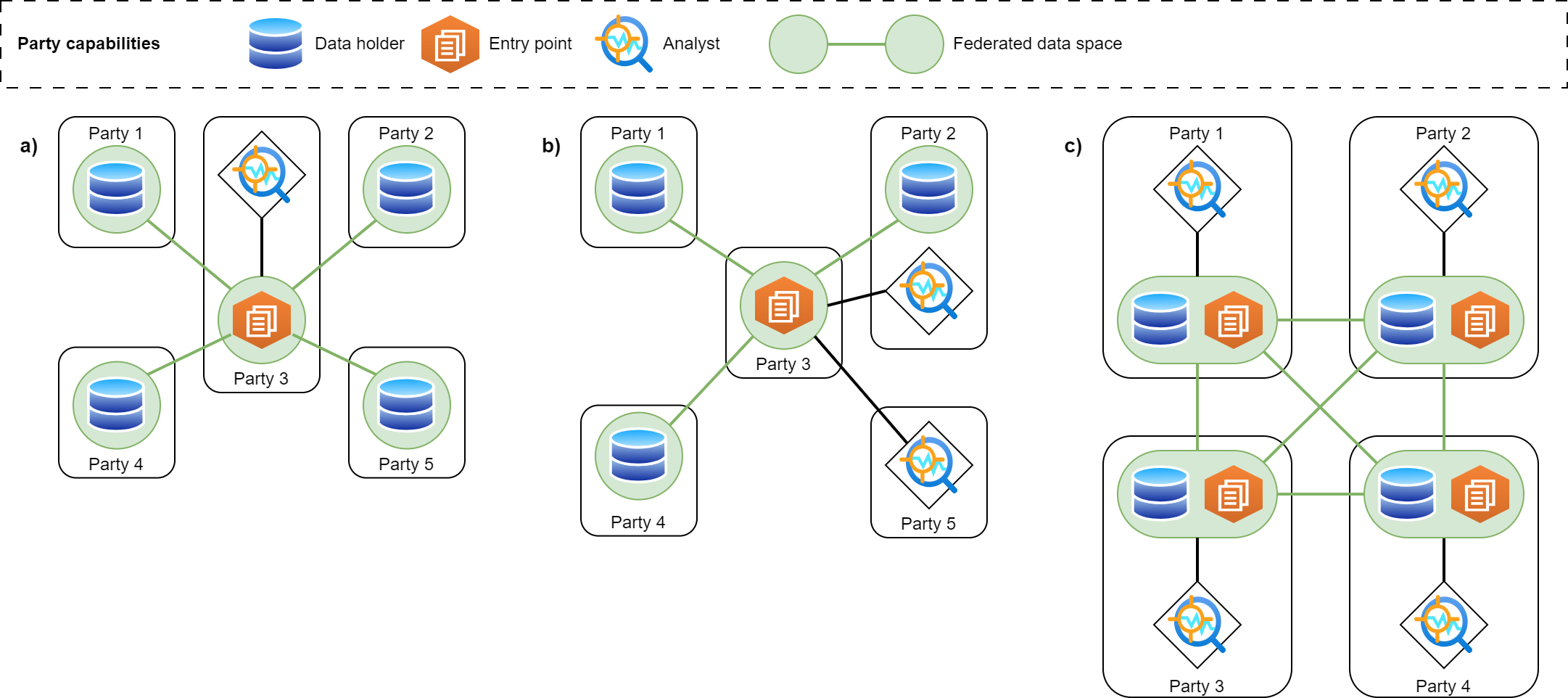}
    \caption{
    Three different deployment topologies that can be realized with \FLib. Analysts can access the federated data space, created by data holders, through one of the entry points of the space.
    Panel \textbf{a)} represents the classic centralized  scenario, in which party 3 acts as the entry point for its own analysts. 
    Panel \textbf{b)} represents a scenario in which analysts are external to the entry point party. 
    Panel \textbf{c)} represents a decentralized scenario, in which all the parties are equivalent and any of them can act as an entry point. 
    }\label{fig:scheme}
\end{figure*}

In the research domain, particularly in healthcare-related initiatives, it is a common situation for multiple parties to engage in a consortium with the aim of developing federated machine learning models using data located at several different sites. This is the case of a number of projects operating within national and international research programs (e.g., Horizon Europe). In such consortia, several parties provide data (\textbf{data holders}), and one party, often not holding any data, conducts the analyses (\textbf{analyst}). Usually, this party also provides an aggregator server that receives model contributions from the other parties involved and builds a global model, thus playing a coordination role that entails both high-level decision-making regarding analyses and low-level task assignments to the different machines involved in the computational process. This scenario follows the classic centralized-topology FL setup, as illustrated in Figure~\ref{fig:scheme}a.

In some cases, consortia can evolve into the creation of a true \textit{federated data space}, where data holders offer the possibility to use their data to any authorized party interested in conducting analysis, possibly including data holders themselves. Figure~\ref{fig:scheme}b depicts this scenario where authorized parties, both holding and not holding data, have the opportunity to conduct analyses. \FLib was developed to address such situations, which demand moving beyond the standard static centralized topology approach. It achieves this by disentangling the aspects of high-level decision-making regarding the kind of analyses to be conducted (analyst capability) from the low-level task assignments (\textbf{entry point} capability, emphasizing with this name the access to the federated data space). In this way, it can naturally handle a wide range of topologies, including cases where all parties are data holders and have the capability of acting both as analysts and as entry points, as shown in Figure~\ref{fig:scheme}c.

Flotta was developed specifically to address the needs of consortia conducting biomedical research. It shares similarities with FeatureCloud \cite{matschinske2023featurecloud}, a ready-to-use solution accessible via a user-friendly web interface which provides algorithms ranging from linear regression to genomic-specific ones, while also allowing custom algorithm development. Both \FLib and FeatureCloud adopt a modular approach, offering a set of secure operations called Apps in FeatureCloud and instructions in \FLib. However, they differ in several key aspects. The main difference concerns the infrastructure; while FeatureCloud requires the use of its own servers, which is often not possible due to parties' internal regulations prohibiting reliance on external servers, \FLib allows relying solely on machines internal to the consortium. Moreover, \FLib was designed to operate in a space where research can be conducted on a number of ready-to-use data resources shared by different parties, enabling authorized analysts to use these resources to perform multiple analyses (limited, for security reasons, to those recognized by the framework). This differs from FeatureCloud, which requires authorization from the data holders for each analysis.

\section{\FLib}
\label{sec:library}

In this section, we move from the high-level description of a consortium based on the functional capabilities of its parties to a process-level description of the functioning of \FLib. The main components for understanding the functioning of \FLib are the Nodes, that contribute to the process with data and computations, the Workbench, where an analyst designs an analysis, and the Artifact, which is created by the Workbench and encodes all the information necessary to run the analysis. We will refer to the set of Nodes as the \textbf{federated data space}: the Workbench is outside of this space and interacts with it by submitting Artifacts to an entry point node of the space.

In the following subsection we provide a more in-depth description of Nodes, Workbenches, and Artifacts. Next, we briefly illustrate the functioning of the library with an example describing the execution of the process depicted in Figure~\ref{fig:workflow}.

\subsection{Components} 
\label{subsec:components}

\textbf{Nodes} are the backbone of any process run through \FLib, as they are deployed for storing local data and carrying out all computations, such as, but not limited to, training machine learning models or aggregating them. Nodes can be characterized from two different perspectives: the functional one and the communication one.

From a functional point of view, Nodes have two main features: scheduling tasks and executing tasks. In a  process only one Node (i.e., the entry point node) is in charge of acting as scheduler. This Node translates the content of the Artifact into a graph of tasks, and distributes these tasks to the other Nodes, following the graph order. The scheduler also tracks the life cycle of all tasks created from an Artifact based on the updates on their status received from the executor Nodes. For example, it unlocks a task that is deeper in the execution graph only once all the tasks on which it depends are successfully completed. The scheduler may also interrupt the process in case of errors. It is worth noting that the scheduler functionality does not exclude the execution of tasks, in fact the entry point node could assign tasks to itself. Moreover, model aggregation is not a special task requiring a dedicated software in our framework; instead, it is a task like any other that the entry point node can assign either to itself or to another Node.

From a communication point of view, Nodes can work in two different modes. The default mode allows to receive communications directly from any other Node. This creates a bidirectional communication channel between two Nodes.
The \textit{client-mode}, in contrast, is a special restricted mode where a Node communicates unidirectionally with only a single reference Node (in most cases, this is the entry point node). When in \textit{client-mode}, a Node cannot be reached from the internet or from a Workbench, and its scheduler functionality is disabled. Instead, at startup, it activates a polling routine that periodically sends requests for updates and tasks to the reference Node. This mode is particularly suited for deploying \FLib with minimal maintenance efforts and in environments where access from the internet is not allowed, such as hospitals.

The Node is implemented reflecting the two previously-explained functional features. It is composed by i) a web service, written using the FastAPI package \cite{Ramirez_FastAPI}, which performs the scheduling of tasks and also permits the interaction with other Nodes and Workbenches, and  ii) the task executor, written using the Ray package \cite{moritz2018ray}, which spawns multiple processes and executes them in parallel. At implementation level, the exchanges of information between Nodes happens thanks to the Pydantic \cite{pydantic} package, which converts Python objects in JSON objects, and back. Exchange of other resources, such as model parameters, instead, is performed through binary data. To enhance the security level of the communication, all the exchanged objects are encrypted using {the asymmetric encryption (or public-key) algorithm}. These keys are also used to identify Nodes and  Workbenches through a signing system; a check on the signature of the transmitted objects allows accepting or discarding requests.

The \textbf{Workbench} is the interface through which an analysis is submitted to the federated data space. Specifically, it is a Python module of \FLib used by the analyst to write the code of the analysis to be executed. In fact, the idea of this interface is inspired by Apache Spark, which offers an interactive interface in Python through the Pyspark package. Analogously, \FLib offers the Workbench interface that allows the interaction with the federated data space. 

We refer to the set of instructions sent from a Workbench to an entry point node as an \textbf{Artifact}. These instructions are also inspired by Apache Spark functions, which are converted and sent to cluster worker nodes before  code  execution. A similar mechanism occurs in \FLib when an Artifact is submitted to the federated data space. Artifacts also share similarities with Extract-Transform-Load pipelines, with the difference that for Artifacts, the steps composing the pipeline are executed at different locations.

A single Artifact is composed of possibly nested multiple instructions, as those defined in the Workbench code example in Listing~\ref{code:example}, that will be discussed in the next subsection. Instructions cover a wide spectrum of operations, ranging from data pre-processing to local model training and model aggregation. As already noted, only a predefined collection of secure instructions can be executed, ensuring the security of the overall analysis. Instructions represent a good trade-off between flexibility and security, offering analysts a wide range of possible operations (to be constructed by composing instructions in different ways), while also preventing the execution of arbitrary code, which could compromise privacy by disclosing sensitive information from the data. 
Moreover, the framework can  be naturally extended by writing custom instructions and wrapping them into plugins. However, in this case, \FLib cannot guarantee the security of the operations executed, and the developer must ensure that the newly defined instructions are indeed secure.

Artifacts are submitted by a Workbench to an entry point node, where each instruction contained in the Artifact is converted into tasks. It is worth noting that a single instruction in an Artifact can generate more than one task: this is the case, for example, of an instruction activating a parallel operation, which generates a task for each of the data holder Nodes involved.

\subsection{Example of process execution}
\label{subsec:process}

\begin{figure}[t]
    \centering
    \includegraphics[width=\linewidth]{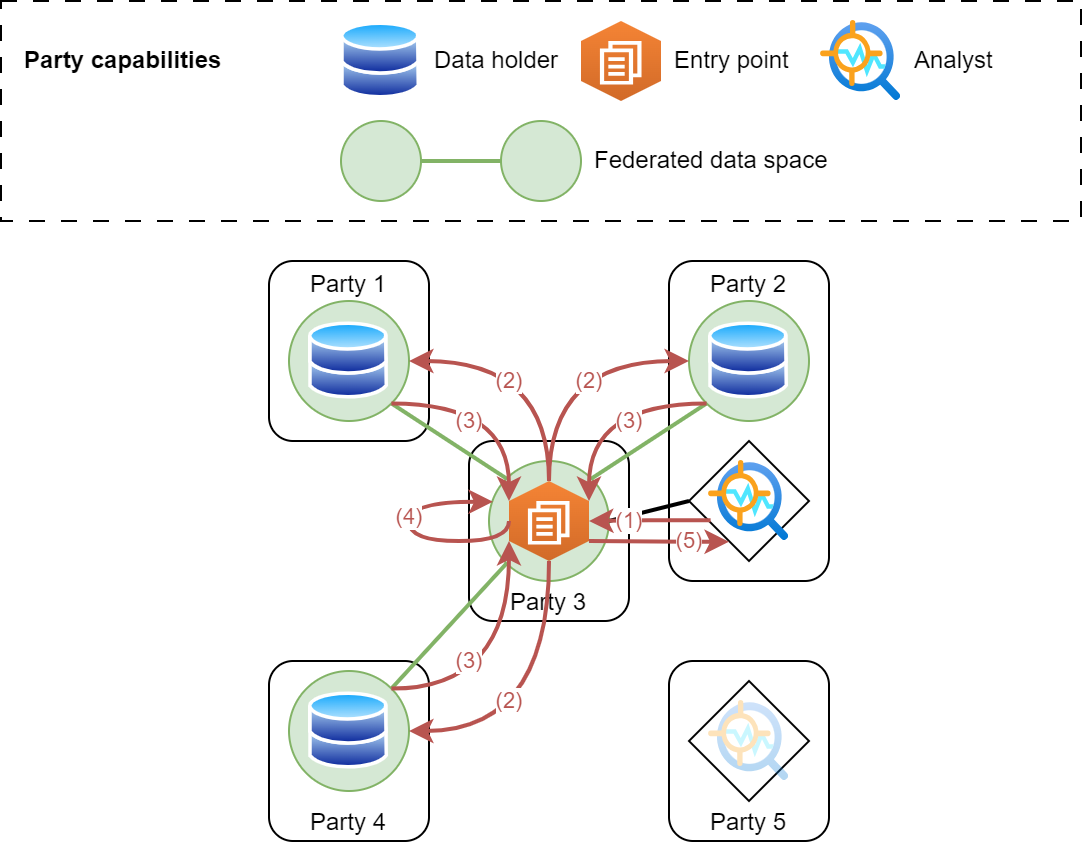}
    \caption{
    Visual representation of the example process outlined in Section~\ref{subsec:process}.
    Arrows in red show the direction of exchanges: Artifact submission (1), tasks distribution (2), local models transmission, (3), aggregation (4), and returning of the global model back to the Analyst (5). 
    In this example, the Analyst of Party 2 is the only analyst working with the federated data space, while the (Analyst of) Party 5 is idle. This can happen as the Analysts of Party 2 and 5 are independent and do not necessarily perform their different analysis at the same time. 
    }\label{fig:workflow}
\end{figure}

In this example, illustrated in Figure~\ref{fig:workflow}, we consider the execution of a process within a five-party consortium. Two parties act as data holders, one acts as both data holder and analyst, one serves as entry point, and one (having only the analyst capability) is inactive. Note that the figure represents the same consortium as in Figure~\ref{fig:scheme}b, this time connoted into a specific process. As our present focus is on the functioning of \FLib, we assume that all the necessary installations and deployments have already been completed (details on this topic are in Section~\ref{sec:usage}), and that all parties are ready to start the process. Specifically, we consider the training of a global random forest model by merging locally trained random forests, which is one of the first experiments that we conducted in the early testing phase of \FLib. In Listing~\ref{code:example}, we provide the code that the Party 2 analyst would need to write in the Workbench for the analysis represented in Figure ~\ref{fig:workflow}. The numbers from (1) to (5) on the arrows of figure \ref{fig:workflow}, which represent the different exchanges of information , are repeated as comments in Listing \ref{code:example} to highlight which operations the different lines of code determine. We first refer to the figure and then to the code, but it may be helpful to examine them both in parallel while reading.

Once the code in Listing~\ref{code:example} is submitted by Party 2, the Workbench generates an Artifact and sends it to the entry point node (1). The entry point node activates a scheduler, creates a graph of tasks, and assigns tasks to the data holder Nodes (2). These locally train their model, and once they have completed the training, send it to the entry point node (3). This latter Node assigns to itself (4) the task of aggregating the local models into a global one, completes the task, and finally sends the global model to the analyst (5). 

\ImportCode

In the code example, lines 1-6 are dedicated to importing the modules that are necessary for executing the process. Line 8 creates a connection between the Workbench and an entry point node; in \FLib, we call the object used by the Workbench to communicate with the entry point node a \textit{context}. Subsequently, it is necessary to fetch a project, which represents a collection of datasets, using a project code (line 9). Lines 11-14 define the machine learning model to be trained by specifying the type of algorithm, one hyperparameter, and the aggregation strategy. Lines 16-34 contain the instructions to train the model inside the federated data space. Instructions are colored in blue in the code and, in this example, are nested. The block of code starting with the \textit{Parallel} instruction at line 17 instructs the scheduler to create multiple tasks that will be executed in parallel on all data holder Nodes. In these parallel operations, each Node, after extracting the data assigned to the project (line 19),  splits the local data into train and test sets (line 20), and trains the model (line 26). Then, line 29 instructs for the collection of the trained local models for the merge step, concluding the operations to be performed in parallel. The block starting at line 31 finalizes the process and therefore requires that all the previous tasks have been successfully completed before it can be executed. In particular, line 32 invokes aggregation using the previously specified aggregation strategy. Finally, line 36 submits the instructions, creating the Artifact and starting the actual execution of the process, while line 37 fetches the global model.

\section{How to use \FLib}
\label{sec:usage}

\FLib is a single, configurable, Python application, distributed as a Python package that contains everything needed to launch a Node application or write code within a Workbench.
It is available as an open source software under the \textit{LGPL 3.0} license. The code is available at the following GitHub repository: \GitHub  while a more in depth documentation of the software is available at \Docs.

In this section, we detail all the necessary steps that a consortium must perform to conduct an analysis with \FLib, starting from the preliminary steps and then detailing the necessary deployments and installations. 

\subsection{Preliminary preparations}
\label{subsec:preparations}

The first step for the consortium is obtaining all the necessary authorizations and approvals for deploying the federated data space and conducting analysis on the data. Even though the federated analysis will preserve privacy, given the relative novelty of the technology, this process may be time-consuming and complicated, especially in biomedical research. Once this phase is concluded, the consortium needs to define the functionalities that each party will have in the federated data space. For instance, due to stringent internal regulations, some parties may only be able to deploy Nodes in client-mode, and therefore only behave as data holders (and analysts). On the other hand, at least one party in the consortium needs to deploy an entry point Node.

Next, the consortium needs to define one or more \textbf{projects}, where a project represents a collection of datasets. Each project is characterized by a unique \textbf{project code}, which is a simple string. A given project code is assigned to all the locally available data files, named \textbf{data source}, belonging to that project (note that each data source may belong to multiple projects). Projects are a key element for the use of \FLib, as every analysis starts with invoking a project code. By creating multiple projects, the consortium can grant access to different ``regions" of the federated data space, i.e., to different subsets of the available datasets. For example, a bioinformatician working in a general-purpose biomedical consortium may be provided with a project code granting access only to genetic datasets instead of the whole set of available datasets, while a nephrologist may have a code granting access only to kidney-related datasets. Projects constitute a flexible way to manage authorizations and contribute to enforcing security in the analyses. Moreover, projects guarantee that, at execution time, the code generated from an Artifact has the expected data ready to be used. Once the different projects have been defined, their codes can be distributed to the different analysts.

\subsection{Node side}
\label{subsec:use_node}

The next step for a consortium is the deployment of Nodes. For this,  a \textbf{configuration file} to be used as input must be prepared for every Node. This file is a document in a human-readable format that contains the deployment parameters of the Node, such as the operative mode (default mode or client-mode) and the connection parameters. The list of data sources provided by a data node, together with the projects to which they are assigned are also included in this document. 

The procedure for deploying a Node is the same for all Nodes in the consortium; the difference is only in the configuration files used. The first Node needs to be configured to operate in the default mode and to be reachable by any other Node in the federated data space (\textbf{cold start} scenario). Once this first Node is added, each subsequent Node will be able to join the federated data space by adding to their configuration file the URL connection to the first Node (\textbf{warm start} scenario). 
As previously mentioned, there is no need for any external entity, such as a cloud provider, outside the consortium parties, for \FLib to work as intended.

Deploying a node involves installing the \FLib package in a Python environment. Additionally, if the intention of the consortium is to use third-party plugins, they shall be installed on every Node during deployment, to avoid having different sets of instructions in the Nodes. After deployment, a Node is launched by executing the install Python package with the configuration file as an argument.
This deployment procedure has been designed in analogy with Apache Spark, since a Spark cluster is created by first installing the same software on all nodes and then configuring each single node. \FLib does the same by offering a single installation package that can be customized based on the necessity of the single party.

\subsection{Workbench side}
\label{subsec:use_workbench}

As stated previously, any party in the consortium, be it a data holder or not, can perform analysis and interact with the federated data space through Workbenches. Multiple Workbenches may be active in the consortium and run different analyses at the same time.

A Workbench needs to be installed in a Python environment, alongside with the same extensions installed for the Nodes in the federated data space, but a Workbench does not require to be launched. In fact, the Workbench can just be imported as a Python package in any sort of Python script, notebook, or application that will perform the analyses.
The functioning of  a Workbench has strong analogies with Apache Spark, where it is possible to create a similar context object using the \textit{Pyspark} module connected to a master node. Through this interface, Spark users write code and send instructions to the master node, which schedule the tasks on the distributed network.

\section{Conclusion}
\label{sec:conclusion}

In this article, we presented \FLib, a FL framework designed for (but not limited to) consortia conducting biomedical research under stringent security requirements. The framework is designed to use machines within the consortium, instead of relying on cloud services. Moreover, the framework provides a good balance between flexibility and security, with the latter being guaranteed through the execution of only a predefined set of granted instructions. The foundational components of \FLib have been developed and tested, and we are currently expanding the selection of algorithms available for model training, as well as the set of possible aggregation strategies. Moreover, provided that the developer ensures their security, the framework can be naturally extended by adding custom instructions and wrapping them into plugins. We encourage interested readers to explore the \FLib code and documentation on GitHub, where updates and future extensions will be readily accessible.

\section*{Acknowledgment}

We are grateful to all SPEARHEAD partners for insightful discussions, in particular to Lina Aerts, Eden Sorolla and Julia Bielicki from Universitäts-Kinderspital beider Basel (UKBB) and Angélie Pham from University of Basel Innovation Office.

\bibliography{IEEEabrv,bibliography}

\end{document}